\documentclass[10pt,twocolumn,letterpaper]{article}

 \usepackage{iccv}              
\usepackage{float}
\usepackage[accsupp]{axessibility}  

\newcommand{\dataset}{\mathcal{D}}

\newcommand{\x}{\mathbf{x}}

\newcommand{\y}{\mathbf{y}}

\newcommand{\w}{\mathbf{w}}
\newcommand{\bfo}{\mathbf{o}}

\newcommand{\Loss}{\mathcal{L}}

\usepackage{xspace}
\newcommand{\rankingloss}{\textbf{Mo}re vs. \textbf{Fe}wer (MoFe) ranking loss\xspace}
\newcommand{\rankinglossfull}{More vs. Fewer ranking loss\xspace}
\newcommand{\rankinglossabbr}{MoFe loss\xspace}
\newcommand{\MoFe}{MoFe\xspace}

\newcommand{\modelabbr}{DMoME\xspace}
\newcommand{\network}{DMoME\xspace}

\definecolor{iccvblue}{rgb}{0.21,0.49,0.74}
\usepackage[pagebackref,breaklinks,colorlinks,allcolors=iccvblue]{hyperref}
\usepackage{algorithm}
\usepackage{algpseudocode}
\usepackage{diagbox}
\usepackage{array}
\usepackage{multirow}
\usepackage{colortbl}
\usepackage{xcolor}
\definecolor{lightgreen}{HTML}{c1e3b8}

\def\paperID{6587} 
\def\confName{ICCV}
\def\confYear{2025}

\title{SimMLM: A Simple Framework for Multi-modal Learning with Missing Modality}

\author{
Sijie Li \quad
Chen Chen \textsuperscript{$\dagger$} \quad
Jungong Han\thanks{Corresponding author} \textsuperscript{$\dagger$} \\
School of Computer Science, University of Sheffield, UK\\
{\tt\small \{sli256, chen.chen2, jungong.han\}@sheffield.ac.uk}
}

\begin{document}

\maketitle
\renewcommand{\thefootnote}{\fnsymbol{footnote}}
\footnotetext[2]{\textsuperscript{$\dagger$}These authors contributed equally as senior authors.}

\begin{abstract} 
In this paper, we propose SimMLM, a simple yet powerful framework for multimodal learning with missing modalities. Unlike existing approaches that rely on sophisticated network architectures or complex data imputation techniques, SimMLM provides a generic and effective solution that can adapt to various missing modality scenarios with improved accuracy and robustness. Specifically, SimMLM consists of a generic Dynamic Mixture of Modality Experts (DMoME) architecture, featuring a dynamic, learnable gating mechanism that automatically adjusts each modality’s contribution in both full and partial modality settings. A key innovation of SimMLM is the proposed More vs. Fewer (MoFe) ranking loss, which ensures that task accuracy improves or remains stable as more modalities are made available. This aligns the model with an intuitive principle: removing one or more modalities should not increase accuracy. We validate SimMLM on multimodal medical image segmentation (BraTS 2018) and multimodal classification (UPMC Food-101, avMNIST) tasks, where it consistently surpasses competitive methods, demonstrating superior accuracy, interpretability, robustness, and reliability across both complete and missing modality scenarios at test time. Code is available at \href{https://github.com/LezJ/SimMLM}{https://github.com/LezJ/SimMLM}.
\end{abstract}    
\section{Introduction}
\label{sec:intro}

Multimodal learning aims to build models that can process and integrate information from multiple, complementary modalities such as visual, textual and auditory information, which has been found useful for various tasks such as image-text food classification~\cite{gallo2020image, liang2022expanding}, multimodal medical image segmentation~\cite{nam2024modality, gao2024training} with improved model accuracy. However, in real-world scenarios, it is common for one or more modalities to be missing during deployment due to hardware failures, environmental conditions, or constraints in data collection, budget, and resources. Addressing this test-time missing modality problem is crucial for building robust multimodal models.

One mainstream to address the missing modality problem is data imputation, which focuses on synthesizing the missing modalities from available ones ~\cite{liu2023m3ae, zhao2024dealing} or approximating intermediate missing features~\cite{wang2023multi, zeng2024missing, wang2024incomplete, kim2024missing}. However, ensuring the factuality and fidelity of synthetic data remains a major challenge, as issues such as poor data quality~\cite{qin2023cross, liang2022mind}, data hallucination~\cite{Sun2024-ob}, and vulnerability to adversarial corruption~\cite{hao2024synthetic, singh2024synthetic} have been reported. Moreover, the computational overhead of generative imputation limits its practicality. In parallel, sophisticated frameworks have been developed, which aim to tackle the missing modality by leveraging advanced, complex network architectures~\cite{chen2019robust, zhang2022mmformer, wang2023multi}, and harmonize or align early or late features from different modalities into a unified, shared space. However, such approaches often introduce substantial resource demands. While some are not adaptable to other tasks at all, others constrain the choice of network architectures or require additional modules, such as projection layers, to accommodate diverse or heterogeneous modalities.

In this paper, we introduce SimMLM, a simple yet effective framework for multi-modal learning with missing modalities, \emph{without relying on specific neural network architectures or complex data imputation}. First, to effectively handle varying modality availability, we propose a Dynamic Mixture of Modality Experts (DMoME) architecture, where modality-specific expert networks process available modalities, and a learnable gating network dynamically adjusts their contributions. This design ensures that the model remains effective in both full and partial modality settings while being adaptable to different neural architectures and task domains.

To further improve the model's robustness and reliability against missing modalities, we introduce a novel loss function, which is designed to optimize the loss landscape around the region of missing data.  Inspired by an intuitive principle that the model should not degrade in performance when modalities are increased, this loss function, named More vs. Fewer (MoFe) ranking loss, ensures that as more modalities become available, the model’s accuracy improves or at least remains stable. The MoFe loss encourages the model to compare performance between modality-rich and modality-poor pairs, promoting an inherent ranking that favors configurations with greater modality availability. This structured comparison in together with the dynamic weighting scheme in \modelabbr allows the model to learn consistent and reliable multimodal fusion strategies, improving robustness and adaptability across varying modality availability.

Our key contributions are summaized as follows:
\begin{itemize}
    \item \textbf{Dynamic Mixture of Modality Experts Architecture:} SimMLM introduces the DMoME architecture with a learnable gating mechanism that adaptively adjusts modality contributions based on availability, improving performance in both complete and partial modality settings. This design enables seamless integration with various neural networks, ensuring flexibility and scalability across diverse tasks.
    \item \textbf{\rankinglossfull:} We introduce the novel MoFe ranking loss, which ensures model performance improves or remains stable as additional modalities are incorporated. By enforcing a ranking comparison between modality-rich and modality-poor pairs, MoFe strengthens model robustness and reliability.
    \item \textbf{Comprehensive Evaluation:} We validate SimMLM on multimodal segmentation and classification across three datasets, demonstrating \emph{superior accuracy, interpretability, robustness, reliability, and computational efficiency} over state-of-the-art methods. These results establish SimMLM as a practical and versatile solution for multimodal learning under missing modality conditions.
\end{itemize}

\begin{figure*}[ht]
    \centering
    \includegraphics[width=0.85\linewidth]{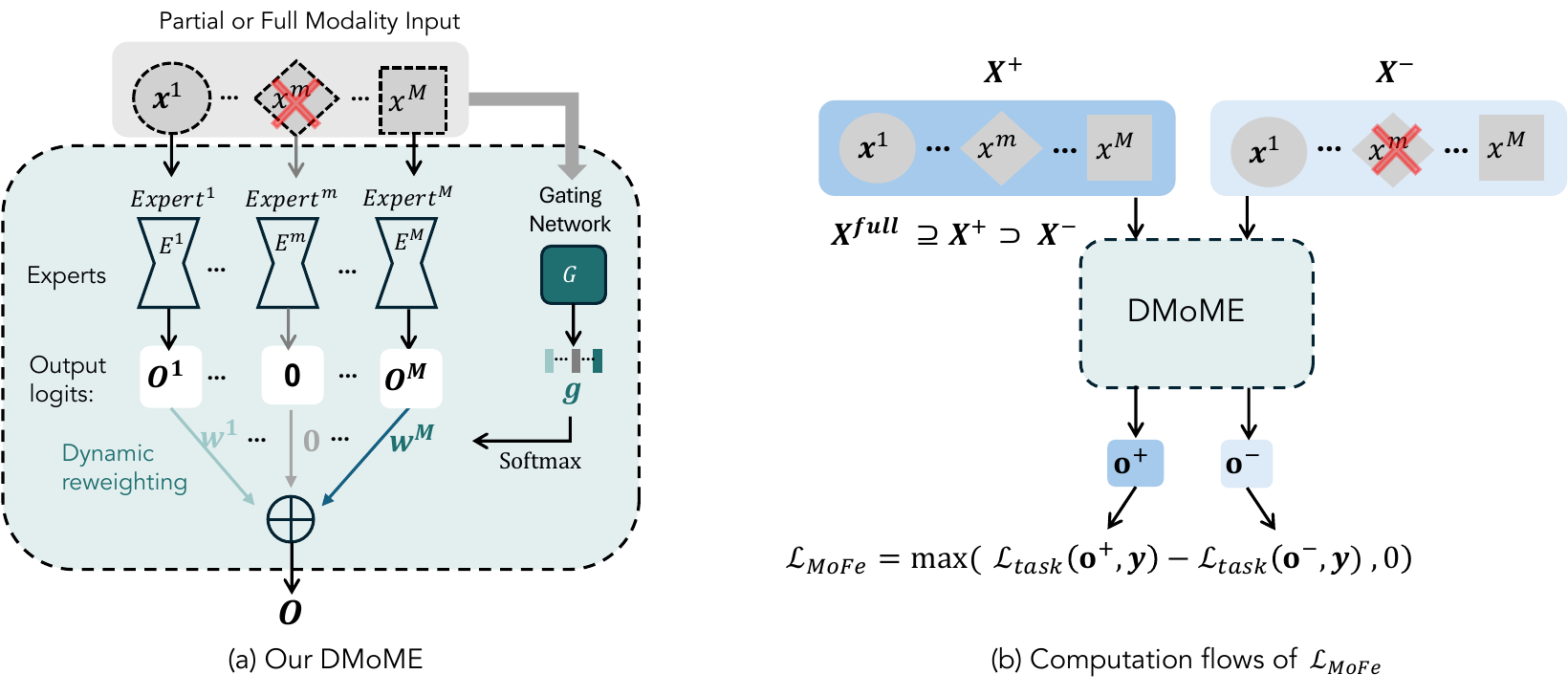}
    \caption{
   Our proposed SimMLM consists of a) DMoME architecture: DMoME takes either partial or full modality inputs where each modality is processed by a corresponding expert network to produce its own output ($   \mathbf{o}^1, \ldots, \mathbf{o}^M$). A gating network (G) assesses the same inputs and produces weights to \emph{dynamically weigh} the contributions of each expert's output in order to get the mixture of output $\mathbf{o}$; b) $\Loss_{\text{MoFe}}$ loss: At training time, in addition to standard task specific losses, DMoME is optimized with the \textbf{Mo}re ($X^+$)  vs. \textbf{Fe}wer Modality ($X^-$) (MoFe) ranking loss, directly encouraging the network to distinguish between inputs with differing modality availability at the task performance level. 
    }
    
    \label{fig: methodology}
\end{figure*}

\section{Related Work}
\label{sec:related_work}
To solve multi-modal learning with missing modalities, various approaches have been proposed, including a) input-level imputation, b) advanced representation learning; and c) mixture of expert strategies~\cite{wu2024deepmultimodallearningmissing}. 

\noindent\textbf{Input-level imputation} approaches leverage advanced generative networks such as auto-encoders~\cite{ma2021smil,liu2023m3ae}, generative adversarial networks~\cite{ma2021smil} to reconstruct missing inputs. Such an approach however, can generate noisy and unreliable input, as discussed in the previous section.

\noindent\textbf{Advanced representation learning} focuses on learning improved representation either through advanced networks such as transformers~\cite{zhang2022mmformer, ma2022multimodal, zhang2024tmformer}, represented by mmFormer~\cite{zhang2022mmformer}, or disentangling the latent features into modality-specific and modality-shared features for improving model accuracy, such as Robust-Mseg~\cite{chen2019robust}and ShaSpec~\cite{wang2023multi}. For instance, ShaSpec~\cite{wang2023multi} employs a combination of modality-shared and modality-specific encoders with dedicated losses to promote feature disentanglement and perform feature fusion in the latent space before passing them to a task decoder. When a modality is missing, ShaSpec generates its features from the available modalities through a missing modality feature generation process enabled by the modality-shared encoder. However, this approach imposes a strict constraint that the shared encoder should be suitable for all input modalities and
both shared and specific features must have the same dimensions, which can be less efficient and optimal when dealing with highly heterogeneous modalities.

\noindent\textbf{Mixture of Experts (MoE)} has recently been adopted in both unimodal and multimodal learning, demonstrating superior performance in natural language processing~\cite{fedus2022switch, zhao2023sparse, lepikhingshard} and computer vision tasks~\cite{shen2024mome, xu2024leveraging, yun2024flex}, while improving efficiency by reducing FLOPS. A very recent method for handling missing modalities is Mixture of Modality Knowledge Experts (MoMKE)~\cite{xu2024leveraging}, which uses MLP experts to generate unimodal and joint representations. Even when a modality is missing, MoMKE leverages existing modalities to generate joint representations by utilizing the other experts. While this approach has achieved promising results, it still incurs high computational costs, as each modality must pass through all modality experts, \emph{quadratically} increasing complexity with the number of modalities. In contrast, the expert model in DMoME is dedicated to a specific modality. Additionally, unlike MoMKE, DMoME does not constrain the dimensionality of latent feature representations across modalities. Instead, it performs the weighting directly at the output space across experts. This flexibility makes DMoME more efficient and scalable, offering high performance in diverse multimodal tasks.

\section{Methodology}

\subsection{Problem Definition}
Assume we have a complete multimodal dataset with $M$ modalities, denoted as $\dataset = \{\{\x_i^m\}_{m=1}^M, \y_i\}_{i=1}^N$. where $\x_i^m \in \mathcal{X}$ represents the $i$-th data sample and $m$ indexes the modality. $\y_i \in \mathcal{Y}$ is the label for the sample $i$ from the label space $\mathcal{Y}$, which can either be a classification label or segmentation map. In this paper, we are interested in finding a solution $f$ that learns a mapping from the input space to the label space: $f:\mathcal{X} \rightarrow \mathcal{Y}$, leveraging this complete modality dataset,  so that $f$ can work robustly and reliably even when certain modalities $m$ are missing at inference time. For simplicity, we omit the sample index $i$ when not specified in the following sections.

\subsection{Overview of SimMLM}
\label{sec: method_overview}
Our proposed SimMLM framework consists of two main components: a) a unified Dynamic Mixture of Modality Experts framework (\modelabbr) (Sec.~\ref{sec: DMoME}), which can take a varied number of modalities as input and adaptively weight their contribution for robust performance at both training and inference time; b) a novel ranking loss called \MoFe (Sec.~\ref{sec: mofe_loss}), which is designed to calibrate the network's performance by differentiating network performance when the number of available modalities for \emph{the same sample} varies.

\subsection{Dynamic Mixture of Modality Expert (DMoME)}
\label{sec: DMoME}
As shown in Figure~\ref{fig: methodology} (a), Our DMoME consists of a set of modality-specific expert networks $\{E^m (\mathbf{x}_m; \mathbf{\theta}_m)\}_{m=1}^{M}$, each containing its own set of learnable parameters $\mathbf{\theta}_m$, and a gating network, $G (\cdot; \mathbf{\phi})$, with a set of learnable parameters $\mathbf{\phi}$. Each expert network, $E^m (\mathbf{x}_m; \mathbf{\theta}_m)$, takes its corresponding modality as input and produces its output in logits $\mathbf{o}^m$, before applying the softmax function:$\bfo^m = E^m(\x^m; \theta^m) \; \in \mathbb{R}^{T}$. Here, $T$ denotes the dimensionality of the output space for a specific task, e.g., the number of predicted classes in a classification task.

The gating network \( G (\cdot; \mathbf{\phi}) \) takes a set of \emph{different modalities} as input and produces a set of gating values \( \{g^m\}_{m=1}^M \) to calibrate different modality experts' output. The \textit{softmax} function is then applied to computing non-negative weights \( \{w_1, w_2, \dots, w_M\} \in \mathbb{R} \), ensuring that the sum of the weights for existing modalities is equal to 1. The final output can be described as a reweighted combination of the outputs from the corresponding modality experts \( \{\mathbf{o}^m\}_{m=1}^M \): $\mathbf{o} = \sum_{m=1}^{M} w^m E^m(\mathbf{x}^m; \theta_m)$.

Notably, the input to the gating network can be an incomplete set of modalities. If certain modalities are missing, those inputs are treated as \( \mathbf{0} \), and the gating values for the corresponding missing modality experts' outputs are set to $ -\infty$ (which causes the corresponding weight values to be 0 after the softmax). Mathematically, the gating outputs are set as:
\begin{equation}
    g_m = \begin{cases} 
          G(\x; \phi)^{(m)}, & \text{if modality } m \text{ is present} \\
      -\infty, & \text{if modality } m \text{ is missing}
   \end{cases},
\end{equation}
where $\x$ can be a complete set of modalities or an arbitrary incomplete set of modalities for a given sample. The final weights therefore can be simply computed as:
{\footnotesize{
\begin{equation}
    w^m = \frac{\exp(g^m)}{\sum_{m'=1}^M \exp(g^{m'})} \quad \text{for} \quad m = 1, 2, \dots, M.
\end{equation}}}
Here, the weight vector \( \w = [w^1, w^2, \dots, w^M] \in \mathbb{R}^{M} \) is non-negative and sums to 1, with \( w^m = 0\) for missing modalities, effectively excluding them from the final mixture. The above configuration can be easily extended to multi-task settings, where each task has its own task-specific weighting scheme. In this case, the gating network produces $\mathbf{w} \in \mathbb{R}^{M \times K}$, with $K$ being the number of tasks.

\subsubsection*{Two-stage training}
To train the whole network, we adopt a two-stage training process, which consists of: 
\begin{enumerate}
    \item \textbf{independent learning}: we independently train a modality-specific expert $E_m$ for each modality $m$, supervised by a task-specific loss $\ell_{\text{task}}(\bfo^m, \y)$. For example, for classification tasks, the loss function can be the cross-entropy (CE) loss, whereas for the segmentation, the loss can be Dice loss. This pretraining stage ensures that each modality expert independently learns specialized features, minimizing the risk of any single modality dominating the learning process. Different modalities can vary significantly in learning dynamics and task difficulty, which can lead to training instability, especially at the early stages of joint training.
    \item \textbf{Cooperative learning}: we co-train the expert networks alongside the gating network, ranking network performance under More vs. Fewer Modality conditions, see next section for details. 
\end{enumerate}
This training scheme enables the expert networks to exploit modality-specific information without interference from other modalities before jointly optimizing with the gating network to exploit mutual benefits. It also improves framework flexibility with reduced training resources, e.g., a) first-stage training can be accelerated via parallel computing; b) when new modalities are introduced, existing pretrained expert networks can be reused for the second stage.

\subsection{More vs. Fewer Modality Co-training} 
\label{sec: mofe_loss}
To boost model robustness against missing modalities, we simulate scenarios with missing modalities by randomly dropping out input modalities during training, which is a common practice. Yet, unlike existing approaches that focus on aligning outputs from missing modalities with those from the full set of modalities~\cite{liu2022moddrop++, hu2020knowledge, karimijafarbigloo2024mmcformer}, we encourage the network to distinguish between modality-rich and modality-poor inputs. \\  
\textbf{Rational behind More vs. Fewer Modality Training: } In many multi-modal learning tasks, all modalities are assumed to be predictive and complementary, contributing valuable information to the tasks that are of interest. Our approach is inspired by the intuition that richer multimodal inputs should yield more accurate results, as each modality complements the others. As the number of modalities decreases, the input becomes more limited, making it more challenging to capture complex patterns. 

As it is difficult to quantify the exact performance for different set-ups, we propose to formulate it as a ranking loss to encourage the network to \emph{compare} the two scenarios.  In cognitive science, it has been found that learners who engage in comparative tasks show an improved ability to transfer knowledge to new situations compared to those who learn through isolated examples~\cite{Tullis2016-fg}. Such a concept has also been applied in reinforcement learning for preference alignment for the applications like large language model training and finetuning with a reward function.

In our multimodal learning setting, at each training iteration, we sample two sets from a data point $\x_{full}$: $\x^{+}$ and ${\x^{-}}$, where ${\x^{-}}\subset{\x^{+}} \subseteq{\x^{full}}$. The corresponding outputs from \network are denoted as $\bfo^{+}$ and ${\bfo^{-}}$, respectively. We conjecture that the task-specific loss value for modality-rich setups with $\x^+$ should be at least no higher than those in the modality $\x^-$. Mathematically,  every pair of
samples $(\x^+,\x^-)$ should satisfy the following relationship:
\[
 \Loss_{task}(\bfo^{-}, \y) \geq  \Loss_{task}(\bfo^{+}, \y) \iff Acc(\x^+)\geq Acc(\x^-).\]
\textbf{Definition of $\Loss_{MoFe}$:} To guide the network to learn the ordinary relationship, we introduce \rankingloss, defined as follows:
\begin{equation}
    \ell_{\text{MoFe}}(\bfo^+, \bfo^-, \y) = \max \left( 0, \ell_{\text{task}}(\bfo^+, \y) - \ell_{\text{task}}(\bfo^-, \y) \right),
\end{equation}
which encourages the network to perform better when more modalities are available. It is worth mentioning that the proposed loss is general and thus can be applied to other existing multi-modal learning frameworks to better calibrate the network for improved robustness against missing modalities as well as improved reliability. 
The total training loss is thus defined as: 
\begin{equation}
   \ell_{\text{total}}(\bfo^+, \bfo^-, \y) = \ell_{\text{task}}(\bfo^+, \y) + \ell_{\text{task}}(\bfo^-, \y)+\lambda   \ell_{\textrm{\MoFe}},
   \label{eq: final_loss}
\end{equation}
where $\lambda$ is a coefficient to control the contribution of $\ell_{\text{MOFE}}$.

We acknowledge the existence of prior research, such as the works of Noh et al. (2023)~\cite{noh2023rankmixup}, and Ma et al. (2023)~\cite{ma2023calibrating}, which have investigated the application of ranking loss at the confidence levels between a pair of two inputs. However, our approach diverges significantly from these existing methods. Specifically, we propose the novel concept of applying ranking loss directly at the loss level, as opposed to the confidence level. This distinction is crucial, as our method aims to \emph{directly regularize the geometry of the loss landscape}.  In our experiments, we also showed extra benefit of our loss, which leads to reduced calibration errors, see Table.~\ref{tab: calibration_error}. Besides, our proposed method enables seamless integration with a range of loss functions, offering flexibility beyond classification and allowing for improved applicability in complex, diverse tasks. For example, our method can be applied for segmentation and regression tasks where the confidence scores are not straightforward in dense or continuous prediction settings. By focusing on the actual task error, our loss-level approach directly aligns with the downstream task performance, providing a more consistent measure across different tasks.

\section{Experiments}
We evaluate our approach on two types of tasks, multimodal medical segmentation and multimodal classification, spanning three datasets. For segmentation, we use BraTS 2018~\cite{menze2014multimodal} benchmark. For classification, we conduct experiments on UPMC Food-101~\cite{gallo2020image} and avMNIST dataset~\cite{vielzeuf2018centralnet}.

\subsection{Datasets}
\noindent \textbf{BraTS 2018} (Brain Tumor Segmentation Challenge~\cite{menze2014multimodal}) is a widely used benchmark for brain tumor segmentation in 3D MRI scans. It has also been extensively employed to evaluate methods for handling missing modalities~\cite{zhang2022mmformer, zhang2024tmformer, wang2023multi, hu2020knowledge}. The dataset includes \emph{four distinct} MRI modalities: FLAIR, T1, T1 contrast-enhanced (T1ce), and T2. The segmentation task involves identifying three tumor subregions: the enhancing tumor (ET), the tumor core (TC), and the whole tumor (WT). The dataset consists of 285 manually labeled cases and 66 unlabeled cases for evaluation. In our experiments, we split the labeled cases into a 4:1 ratio for training and validation. Final evaluations are conducted on the 66 unlabeled cases via the official benchmarking platform for a consistent and fair accuracy comparison against existing methods~\cite{lee2023multimodal, zhang2022mmformer}.

\noindent \textbf{UPMC Food-101}~\cite{gallo2020image} is a large-scale food classification dataset with 90,704 image-text pairs spanning 101 classes. Collected from uncontrolled environments, it introduces inherent noise, posing challenges for accurate classification.

\noindent \textbf{avMNIST} (Audiovision MNIST~\cite{vielzeuf2018centralnet}) is a digit classification dataset comprising 1,500 image-audio pairs. The visual modality consists of $28\times28$ grayscale images of handwritten digits (0-9) from MNIST~\cite{lecun1998gradient}, while the audio modality is sourced from the Free Spoken Digits Dataset\footnote{Dataset link: \href{https://github.com/Jakobovski/free-spoken-digit-dataset}{https://github.com/Jakobovski/free-spoken-digit-dataset}.}.

\subsection{Implementation Details}
\label{sec: implementation}
All the experiments were conducted on a single A100 GPU node. Detailed implementations are described below. 

\noindent \textbf{BraTS 2018:} 
We employed nnUNet~\cite{isensee2021nnu}, a widely used and powerful medical image segmentation backbone, as the expert network for each modality, which can be viewed as a 3D U-Net. The gating network $G$ in \modelabbr utilized lightweight CNNs to capture broad contextual information, followed by a linear layer generating $4 \times 3$ weights for the four modalities and three segmentation tasks. We optimized both training stages using the Adam optimizer~\cite{kingma2014adam} ($lr=0.01$). The $\Loss_{task}$ in Eq.\ref{eq: final_loss} combined Dice~\cite{milletari2016v} and binary cross-entropy losses.

\noindent \textbf{UPMC Food-101:} 
We employed Inception V3~\cite{szegedy2016rethinking} as the image expert and BERT~\cite{devlin2019bert} as the text expert, following~\cite{gallo2020image}. To avoid redundant feature extraction, intermediate features from both were fed into an MLP gating network $G$, which generated 2D weights for expert outputs. Optimization was performed using Adam ($lr=0.0001$) with cross-entropy loss as $\Loss_{task}$.

\noindent \textbf{avMNIST:} 
Following~\cite{ma2021smil, wang2023multi}, we adopted a LeNet-5~\cite{lecun1998gradient} backbone for both image and audio modalities, with audio converted into $20\times20$ MFCC spectrograms~\cite{tzanetakis2002musical}. The gating network $G$ followed a similar design as in BraTS 2018. Two-stage training was conducted using Adam ($lr=0.001$) with cross-entropy loss as $\Loss_{task}$.

Detailed gating network $G$ architectures for BraTS 2018 and avMNIST are provided in the Supplementary. For task accuracy evaluation, we follow common practice, using the Dice score for segmentation (BraTS 2018) and Accuracy for classification (UPMC Food-101, avMNIST). The \rankinglossabbr~coefficient $\lambda$ in Eq.\ref{eq: final_loss} is set to 0.1 across tasks; however, our method remains robust to variations in $\lambda$, as demonstrated in Sec.~\ref{analysis} Analysis.

\subsection{Segmentation Results}

\begin{table*}[t]
    \centering
    \setlength{\tabcolsep}{3pt}
    \footnotesize
\resizebox{0.85\textwidth}{!}{%
\begin{tabular}{@{}llcccc|cccc>{\columncolor{lightgreen}}c|cccc>{\columncolor{lightgreen}}c|cccc>{\columncolor{lightgreen}}c@{}}

\toprule
\multicolumn{2}{l|}{\multirow{2}{*}{Settings}} & \multicolumn{4}{c|}{\begin{tabular}[c]{@{}c@{}}Available Modalities \\ at Test Time\end{tabular}} & \multicolumn{5}{c|}{Enhancing tumor (ET)} & \multicolumn{5}{c|}{Tumor core (TC)} & \multicolumn{5}{c}{Whole tumor (WT)} \\ \cline{3-21} 
\multicolumn{2}{l|}{} &   \rotatebox{45}{Fl} & \rotatebox{45}{T1} & \rotatebox{45}{T1ce} & \rotatebox{45}{T2} & \begin{tabular}[c]{@{}c@{}}RbSeg\\ \cite{chen2019robust}\end{tabular} & \begin{tabular}[c]{@{}c@{}}mmFm\\ \cite{zhang2022mmformer}\end{tabular} & \begin{tabular}[c]{@{}c@{}}ShaSpec \\ \cite{wang2023multi}\end{tabular} & \multicolumn{1}{c|}{\begin{tabular}[c]{@{}c@{}}MoMKE\\ \cite{xu2024leveraging}\end{tabular}} & Ours & \begin{tabular}[c]{@{}c@{}}RbSeg\\ \cite{chen2019robust}\end{tabular} & \begin{tabular}[c]{@{}c@{}}mmFm\\ \cite{zhang2022mmformer}\end{tabular} & \begin{tabular}[c]{@{}c@{}}ShaSpec \\ \cite{wang2023multi}\end{tabular} & \multicolumn{1}{c|}{\begin{tabular}[c]{@{}c@{}}MoMKE\\ \cite{xu2024leveraging}\end{tabular}} & Ours & \begin{tabular}[c]{@{}c@{}}RbSeg\\ \cite{chen2019robust}\end{tabular} & \begin{tabular}[c]{@{}c@{}}mmFm\\ \cite{zhang2022mmformer}\end{tabular} & \begin{tabular}[c]{@{}c@{}}ShaSpec \\ \cite{wang2023multi}\end{tabular} & \multicolumn{1}{c|}{\begin{tabular}[c]{@{}c@{}}MoMKE\\ \cite{xu2024leveraging}\end{tabular}} & Ours \\  \midrule

\multicolumn{1}{l|}{\multirow{14}{*}{\rotatebox{90}{Missing}}} & \multicolumn{1}{l|}{1} & $\bullet$ & $\circ$ & $\circ$ & $\circ$ & 25.69 & 39.33 & 43.52 & \multicolumn{1}{c|}{\underline{50.53}} & \textbf{51.31} & 53.57 & 61.21 & 69.44 & \multicolumn{1}{c|}{\textbf{75.05}} & \underline{74.12} & 85.69 & 86.10 & 88.68 & \multicolumn{1}{c|}{\textbf{89.78}} & \underline{89.24} \\ 
\cmidrule(l){2-21} 
\multicolumn{1}{l|}{} & \multicolumn{1}{l|}{2} & $\circ$ & $\bullet$ & $\circ$ & $\circ$ & 17.29 & 32.53 & 41.00 & \multicolumn{1}{c|}{\underline{48.12}} & \textbf{48.41} & 47.90 & 56.55 & 63.18 & \multicolumn{1}{c|}{\underline{68.66}} & \textbf{71.48} & 70.11 & 67.52 & 73.44 & \multicolumn{1}{c|}{\underline{77.45}} & \textbf{80.28} \\ 
\cmidrule(l){2-21} 
\multicolumn{1}{l|}{} & \multicolumn{1}{l|}{3} & $\circ$ & $\circ$ & $\bullet$ & $\circ$ & 67.07 & 72.60 & \underline{73.29} & \multicolumn{1}{c|}{72.30} & \textbf{78.58} & 76.83 & 75.41 & 78.65 & \multicolumn{1}{c|}{\underline{78.71}} & \textbf{83.45} & 73.31 & 72.22 & 73.82 & \multicolumn{1}{c|}{\underline{74.00}} & \textbf{80.02} \\
\cmidrule(l){2-21} 
\multicolumn{1}{l|}{} & \multicolumn{1}{l|}{4} & $\circ$ & $\circ$ & $\circ$ & $\bullet$ & 28.97 & 43.05 & 46.31 & \multicolumn{1}{c|}{\textbf{51.02}} & \underline{50.60} & 57.49 & 64.20 & 69.03 & \multicolumn{1}{c|}{\underline{72.34}} & \textbf{73.49} & 82.24 & 81.15 & 83.99 & \multicolumn{1}{c|}{\underline{85.82}} & \textbf{86.01} \\ 
\cmidrule(l){2-21} 
\multicolumn{1}{l|}{} & \multicolumn{1}{l|}{5} & $\bullet$ & $\bullet$ & $\circ$ & $\circ$ & 32.13 & 42.96 & 44.76 & \multicolumn{1}{c|}{\underline{52.52}} & \textbf{53.62} & 60.68 & 65.91 & 72.67 & \multicolumn{1}{c|}{\underline{75.00}} & \textbf{77.14} & 88.24 & 87.06 & 89.76 & \multicolumn{1}{c|}{\textbf{90.23}} & \underline{89.98} \\ 
\cmidrule(l){2-21} 
\multicolumn{1}{l|}{} & \multicolumn{1}{l|}{6} & $\bullet$ & $\circ$ & $\bullet$ & $\circ$ & 70.30 & 75.07 & 75.60 & \multicolumn{1}{c|}{\underline{78.75}} & \textbf{80.25} & 80.62 & 77.88 & 84.50 & \multicolumn{1}{c|}{\underline{84.62}} & \textbf{84.96} & 88.51 & 87.30 & 90.06 & \multicolumn{1}{c|}{\underline{90.34}} & \textbf{90.35} \\ 
\cmidrule(l){2-21} 
\multicolumn{1}{l|}{} & \multicolumn{1}{l|}{7} & $\bullet$ & $\circ$ & $\circ$ & $\bullet$ & 33.84 & 47.52 & 47.20 & \multicolumn{1}{c|}{\underline{54.12}} & \textbf{54.62} & 61.16 & 69.75 & 72.93 & \multicolumn{1}{c|}{\textbf{76.90}} & \underline{75.76} & 88.28 & 87.59 & 90.02 & \multicolumn{1}{c|}{\textbf{90.53}} & \underline{90.28} \\ 
\cmidrule(l){2-21} 
\multicolumn{1}{l|}{} & \multicolumn{1}{l|}{8} & $\circ$ & $\bullet$ & $\bullet$ & $\circ$ & 69.06 & 74.04 & 75.76 & \multicolumn{1}{c|}{\underline{76.78}} & \textbf{78.77} & 78.72 & 78.59 & \underline{82.10} & \multicolumn{1}{c|}{81.70} & \textbf{84.79} & 77.18 & 74.42 & 78.74 & \multicolumn{1}{c|}{\underline{79.86}} & \textbf{82.25} \\ 
\cmidrule(l){2-21} 
\multicolumn{1}{l|}{} & \multicolumn{1}{l|}{9} & $\circ$ & $\bullet$ & $\circ$ & $\bullet$ & 32.01 & 44.99 & 46.84 & \multicolumn{1}{c|}{\textbf{53.74}} & \underline{53.40} & 62.19 & 69.42 & 71.38 & \multicolumn{1}{c|}{\underline{73.73}} & \textbf{74.73} & 84.78 & 82.20 & 86.03 & \multicolumn{1}{c|}{\underline{86.60}} & \textbf{87.42} \\ 
\cmidrule(l){2-21} 
\multicolumn{1}{l|}{} & \multicolumn{1}{l|}{10} & $\circ$ & $\circ$ & $\bullet$ & $\bullet$ & 69.71 & 74.51 & 75.95 & \multicolumn{1}{c|}{\underline{77.11}} & \textbf{78.61} & 80.20 & 78.61 & \underline{83.82} & \multicolumn{1}{c|}{83.61} & \textbf{84.19} & 85.19 & 82.99 & 85.42 & \multicolumn{1}{c|}{\underline{86.92}} & \textbf{88.11} \\ 
\cmidrule(l){2-21} 
\multicolumn{1}{l|}{} & \multicolumn{1}{l|}{11} & $\bullet$ & $\bullet$ & $\bullet$ & $\circ$ & 70.78 & 75.47 & 76.42 & \multicolumn{1}{c|}{\underline{78.19}} & \textbf{80.41} & 81.06 & 79.80 & \underline{85.23} & \multicolumn{1}{c|}{82.83} & \textbf{85.25} & 88.73 & 87.33 & \underline{90.29} & \multicolumn{1}{c|}{90.21} & \textbf{90.63} \\ 
\cmidrule(l){2-21} 
\multicolumn{1}{l|}{} & \multicolumn{1}{l|}{12} & $\bullet$ & $\bullet$ & $\circ$ & $\bullet$ & 36.41 & 47.70 & 46.55 & \multicolumn{1}{c|}{\textbf{55.93}} & \underline{55.84} & 64.38 & 71.52 & 73.97 & \multicolumn{1}{c|}{\underline{77.00}} & \textbf{78.02} & 88.81 & 87.75 & 90.36 & \multicolumn{1}{c|}{\underline{90.59}} & \textbf{90.62} \\ 
\cmidrule(l){2-21} 
\multicolumn{1}{l|}{} & \multicolumn{1}{l|}{13} & $\bullet$ & $\circ$ & $\bullet$ & $\bullet$ & 70.88 & 75.67 & 75.99 & \multicolumn{1}{c|}{\underline{78.44}} & \textbf{80.29} & 80.72 & 79.55 & \textbf{85.26} & \multicolumn{1}{c|}{83.59} & \underline{85.23} & 89.27 & 88.14 & 90.78 & \multicolumn{1}{c|}{\underline{90.85}} & \textbf{90.99} \\ 
\cmidrule(l){2-21} 
\multicolumn{1}{l|}{} & \multicolumn{1}{l|}{14} & $\circ$ & $\bullet$ & $\bullet$ & $\bullet$ & 70.10 & 74.75 & 76.37 & \multicolumn{1}{c|}{\underline{78.27}} & \textbf{80.33} & 80.33 & 80.39 & \underline{84.18} & \multicolumn{1}{c|}{82.35} & \textbf{85.08} & 86.01 & 82.71 & 86.47 & \multicolumn{1}{c|}{\underline{86.62}} & \textbf{87.78} \\ 
\midrule
\multicolumn{1}{l|}{Full} & \multicolumn{1}{l|}{15} & $\bullet$ & $\bullet$ & $\bullet$ & $\bullet$ & 71.13 & 77.61 & \underline{78.08} & \multicolumn{1}{c|}{77.65} & \textbf{80.40} & 80.86 & \textbf{85.78} & \underline{85.45} & \multicolumn{1}{c|}{82.61} & 85.39 & 89.45 & 89.64 & \underline{90.88} & \multicolumn{1}{c|}{90.54} & \textbf{91.07} \\ 
\midrule
\multicolumn{6}{c|}{Average} & 51.02 & 59.85 & 61.58 & \multicolumn{1}{c|}{\underline{65.56}} & \textbf{67.16} & 69.78 & 72.97 & 77.45 & \multicolumn{1}{c|}{\underline{78.58}} & \textbf{80.20} & 84.39 & 82.94 & 85.92 & \multicolumn{1}{c|}{\underline{86.69}} & \textbf{87.67} \\ \bottomrule
\end{tabular}%
}
    \caption{Model accuracy (\textbf{Dice score $\uparrow$} for the segmentation task, normalized to 100\%) comparison of different methods for segmenting brain tumor subregions under various missing ($\circ$) and/or full modality settings. The best and second best performance of each subtask for each setting are highlighted in \textbf{bold} and \underline{underline} respectively.}
    \label{tab: seg_res}
\end{table*}

\textbf{Benchmarking segmentation accuracy on the official BraTS 2018 evaluation set.} For the BraTS 2018 benchmark, we compare our approach against several state-of-the-art methods, including Robust-MSeg (RbSeg)~\cite{chen2019robust}, mmFormer (mmFm)~\cite{zhang2022mmformer}, ShaSpec~\cite{wang2023multi}, and MoMKE~\cite{xu2024leveraging} under 15 configurations of both missing and full modality set-ups in Table~\ref{tab: seg_res}. Results show that SimMLM achieves the highest average accuracy in terms of the Dice scores, consistently ranking as the top or second-highest performer in each entry across 44 out of all 45 configurations (15 settings $\times$ 3 subtasks), despite utilizing only 80\% of the labeled data compared to the 100\% used by ShaSpec and mmFormer. This consistent performance highlights our model's robustness and data efficiency in handling various incomplete modality inputs. Notably, when only the T1ce modality is available (Table~\ref{tab: seg_res}, row 3), our model significantly improves the Dice score compared to the second-best methods, increasing from 73.29 to 78.58 for ET, 78.71 to 83.45 for TC, and 74.00 to 80.02 for WT. Such results underscore the advantage of our approach even in single-modality settings.

\begin{figure*}[t]
    \centering
    \includegraphics[width=0.83\textwidth]{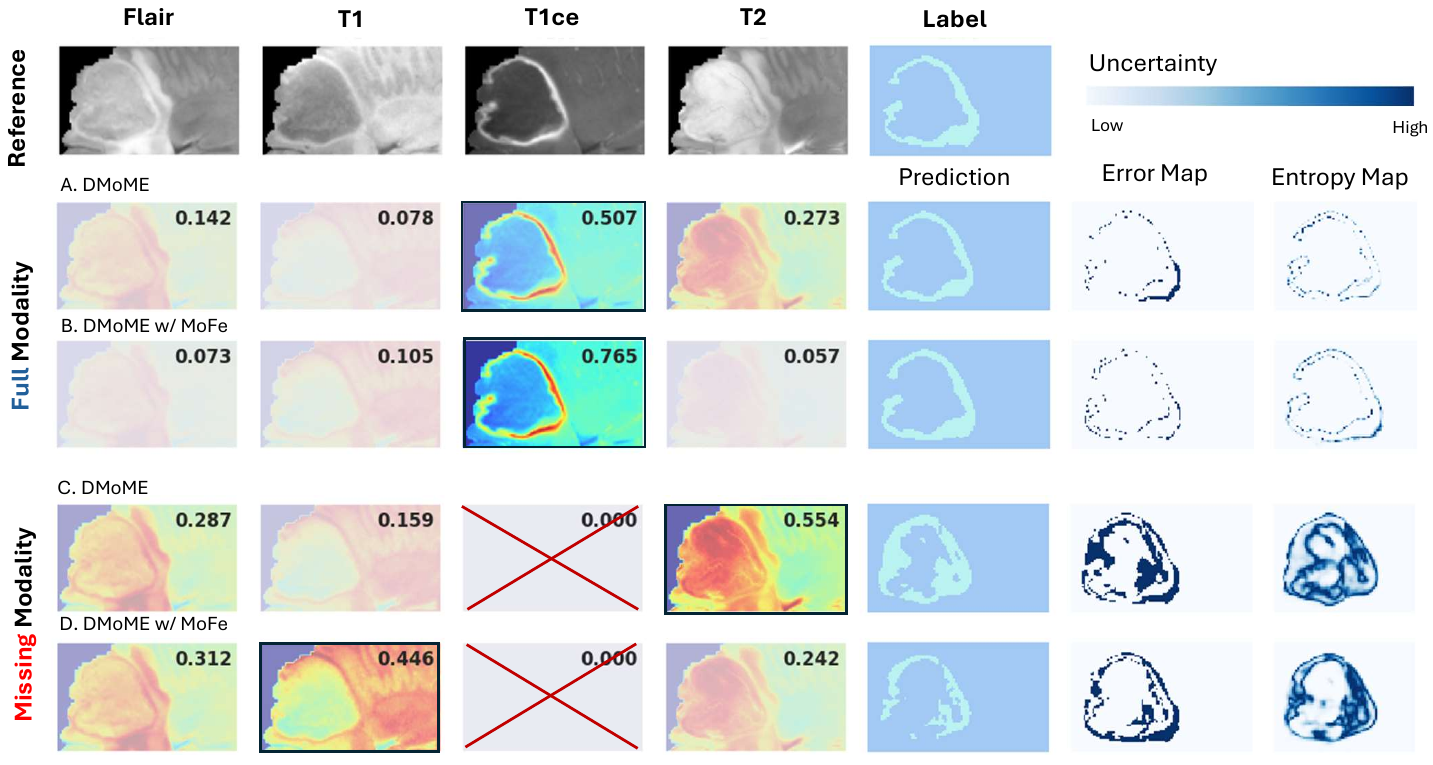} 
    \caption{Visualization of modality weight changes (\textbf{bold numbers} in the upper right corner) assigned by the gating network alongside the predicted ET segmentation and uncertainty (entropy) maps when full or partial modalities (T1ce is missing) are present. DMoME with $\Loss_{MoFe}$ can intelligently shift the focus to the most relevant modality: T1 (the unenhanced counterpart of T1ce) when T1ce is missing, which is aligned with clinical practice despite the network is not explicitly taught.}
    \label{fig: case_study}
\end{figure*}

As shown in Figure~\ref{fig: case_study}, in both full and partial modality scenarios, SimMLM (DMoME with $\Loss_{MoFe}$) achieves more accurate segmentation results with better model reliability compared to DMoME alone (A vs. B, C vs. D) thanks to the contribution of $\Loss_{MoFe}$.  When T1ce is unavailable, DMoME with $\Loss_{MoFe}$ can intelligently shift the focus to the most relevant modality: T1 (the unenhanced counterpart of T1ce), which is aligned with clinical practice despite the network is not explicitly taught.  Without MoFe loss, the network is biased to T2 (maybe due to higher intensity) and \emph{over-segment} the tumor. In the following Sec.~\ref{sec:analysis}, we will provide a more detailed quantitative analysis for our method.  

\subsection{Classification Results}

\begin{table}[t]
    \centering
    \setlength{\tabcolsep}{3pt}
\centering
\resizebox{0.9\columnwidth}{!}{%
\begin{tabular}{@{}c|cc|cc>{\columncolor{lightgreen}}c@{}}
\toprule
\textbf{Dataset} & \multicolumn{2}{c|}{\textbf{Modalities}} & \multicolumn{3}{c}{\textbf{Accuracy (\%) $\uparrow$}} \\ 
\cmidrule(lr){2-3} \cmidrule(lr){4-6} 
& Modality 1 & Modality 2 & \multicolumn{1}{c|}{ShaSpec~\cite{wang2023multi}} & \multicolumn{1}{c|}{MoMKE~\cite{xu2024leveraging}} & SimMLM (Ours) \\ 
\midrule
\midrule

 \multirow{4}{*}{\textbf{UPMC Food-101}} & Image & $\circ$ & \multicolumn{1}{c|}{69.22} & \multicolumn{1}{c|}{70.46} & \textbf{72.20}  \\ 
 & $\circ$ & Text & \multicolumn{1}{c|}{86.55} & \multicolumn{1}{c|}{86.59} & \textbf{87.2}  \\ 
 & Image & Text & \multicolumn{1}{c|}{92.73} & \multicolumn{1}{c|}{92.71} & \textbf{94.99} \\ 
\cmidrule(lr){2-6}
 & \multicolumn{2}{c|}{\textbf{Average}} & \multicolumn{1}{c|}{82.83} & \multicolumn{1}{c|}{83.25} & \textbf{84.81} \\ 
 
\midrule
\midrule

\multirow{4}{*}{\textbf{avMNIST}} & Image & $\circ$ & \multicolumn{1}{c|}{91.90} & \multicolumn{1}{c|}{92.61} & \textbf{92.69} \\ 
 & $\circ$ & Audio & \multicolumn{1}{c|}{89.28} & \multicolumn{1}{c|}{91.16} & \textbf{91.61} \\ 
 & Image & Audio & \multicolumn{1}{c|}{98.71} & \multicolumn{1}{c|}{98.69} & \textbf{99.27}\\ 
\cmidrule(lr){2-6}
 & \multicolumn{2}{c|}{\textbf{Average}} & \multicolumn{1}{c|}{93.30} & \multicolumn{1}{c|}{94.15} & \textbf{94.52} \\ 
 
\bottomrule
\end{tabular}%
}
    \caption{\textbf{Accuracy (\%) $\uparrow$} comparison across missing \& full modality settings on UPMC Food-101 and avMNIST datasets. The datasets have different modality pairs (Image-Text for UPMC Food-101, Image-Audio for avMNIST). $\circ$ represents modality missingness.}
    \label{tab: cls_res}
\end{table}

\textbf{Benchmarking classification accuracy on UPMC Food-101 and avMNIST.} For the classification tasks, we compared our model to the two most competitive methods: ShaSpec~\cite{wang2023multi} and MoMKE~\cite{xu2024leveraging}. To ensure fair comparisons, all models were trained under identical conditions. The results, presented in Table~\ref{tab: cls_res}, report accuracy across different modality settings. Our method consistently surpasses SOTA methods across all settings, demonstrating its effectiveness in handling both missing and full modality scenarios. While clear improvements are observed on both datasets, our method achieves even greater gains on the \emph{noisy} UPMC Food-101 dataset, where classification is more challenging, underscoring the robustness of our SimMLM.

\section{Analysis}
\label{analysis}
\label{sec:analysis}
\textbf{Effectiveness of \modelabbr:}
\begin{table}[t]
    \centering
    \resizebox{0.9\columnwidth}{!}{%
    \begin{tabular}{l|ccc|cc}
        \toprule
        & \multicolumn{3}{c|}{\textbf{Dice score (\%) $\uparrow$}} & \multicolumn{2}{c}{\textbf{Model efficiency}} \\
        \midrule
        Method & ET & TC & WT & \#Params & FLOPS \\
        \midrule
        mmFormer~\cite{zhang2022mmformer} & 59.85 & 72.97 & 82.94 & 106M & 748G \\
        ShaSpec~\cite{wang2023multi} & 61.58 & 77.45 & 85.92 & 187.7M & 713G \\
        MoMKE~\cite{xu2024leveraging} & 65.56 & 78.58 & 86.69 & \textbf{7.8M} & 490G \\
        \rowcolor{lightgreen}
        \modelabbr w/o $\Loss_{MoFe}$ & \textbf{66.05} & \textbf{79.14} & \textbf{86.71} & \textbf{7.8M} & \textbf{123G} \\
        \midrule
        \modelabbr w $\Loss_{MoFe}$ (SimMLM) & 67.16 & 80.20 & 87.67 & 7.8M & 123G \\
        \bottomrule
    \end{tabular}}
    \caption{Comparison of model accuracy (\textbf{Dice score $\uparrow$}, normalized to 100\%, \emph{averaged across all missing \& full modality settings}) and efficiency ($\#$ of parameters and FLOPS) on BraTS 2018 official-evaluated benchmark. }
    \label{tab: models_comp}
\end{table}
To assess the unique contribution of \modelabbr, we trained the network w/o MoFe loss: $\Loss_{MoFe}$ and  compared its performance with competitive models—mmFormer~\cite{zhang2022mmformer}, ShaSpec~\cite{wang2023multi}, and MoMKE~\cite{xu2024leveraging}. Experiments were conducted on the BraTS 2018 dataset, encompassing 15 different configurations for both missing-modality and full-modality setups. Table~\ref{tab: models_comp} reports the average Dice scores across 15 scenarios, alongside networks' parameters and floating-point operations per second (FLOPS). As a result, \modelabbr achieves the highest Dice scores across all tumor subregions, with particularly strong performance in the more challenging enhancing tumor (ET) and tumor core (TC) segmentation. Importantly, \modelabbr achieves a \emph{fourfold} reduction in FLOPS compared to MoMKE, even with the same expert architecture, and requires \emph{over 13 times} fewer parameters than mmFormer and ShaSpec, underscoring its efficiency and scalability. This reduction in FLOPS is due to a key difference in processing: unlike MoMKE, which feeds each modality's input to all experts ~\cite{xu2024leveraging}, our \modelabbr processes each input with only its dedicated expert. This simple approach significantly reduces FLOPS without compromising representational quality.
Further details on \modelabbr's micro design are provided in the Supplementary.

\vspace{5pt}
\noindent\textbf{Interpretability of DMoME:}
\begin{figure}[t]
    \centering
\includegraphics[width=0.9\columnwidth]{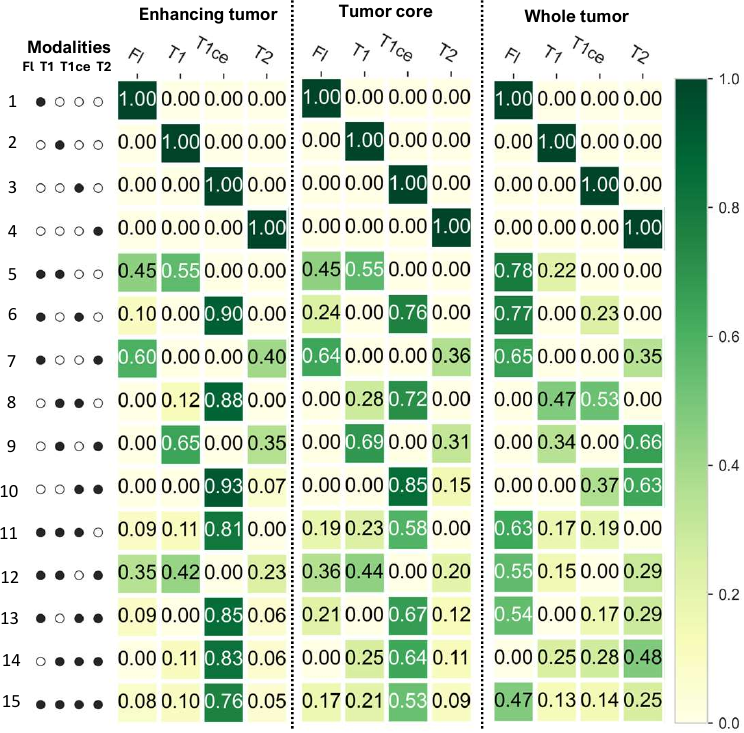}
    \caption{\textbf{Average gating weights analysis} under both full and missing modality setups.  Plotted values are the average gating weights on the BRATS 2018 official evaluation set.}
    \label{fig: gating_weights}
\end{figure}
\begin{figure}[t]
    \centering
    \includegraphics[width=1\columnwidth]{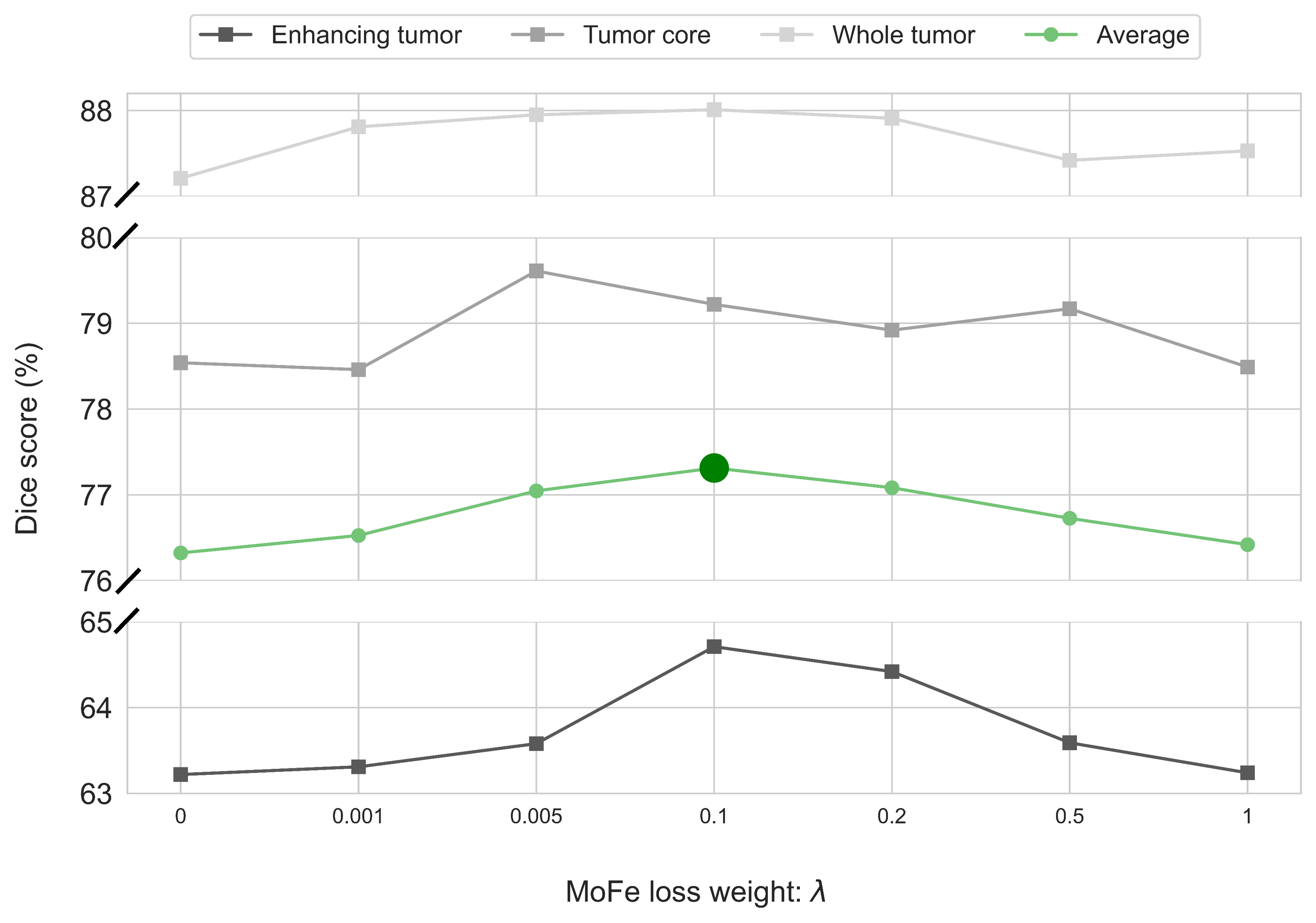} 
    \caption{\textbf{Impact of the MoFe loss coefficient $\lambda$}. The \textbf{Dice score (\%) $\uparrow$} is reported with different values of coefficients $\lambda$ for the MoFe loss across all missing \& full modality settings on the BraTS 2018 validation set. The results include (1) enhancing tumor, (2) tumor core, (3) whole tumor, and (4) their average.}
    \label{fig: lambda_tuning}
\end{figure}
Importantly, our network directly provides \emph{case-specific} interpretability by dynamically revealing the importance of each modality, as shown in  Fig.~\ref{fig: case_study}. In contrast, previous works such as mmFormer~\cite{zhang2022mmformer}, ShaSpec~\cite{wang2023multi}, and MoMKE~\cite{xu2024leveraging} lack this capability. We further performed population-wise study by plotting  the average gating weights under different full- and missing-modality configurations on the official BraTS 2018 evaluation set in Fig.~\ref{fig: gating_weights} across different tasks. It is evident that the gating network managed to dynamically adapt the weights to accommodate various scenarios, providing additional insights for clinicians regarding modality importance in different subtasks. For instance, T1ce is prioritized for enhancing tumor (ET) segmentation, while Flair and T2 are more important for whole tumor (WT) segmentation, which is consistent with clinical findings~\cite{menze2014multimodal}.

\label{sec: mofe_effectiveness}

\vspace{5pt}
\noindent\textbf{Effectiveness and Sensitivity of MoFe loss:} We further analyzed the impact of \rankinglossabbr by varying its coefficient $\lambda = \{0, 0.01, 0.05, 0.1, 0.2, 0.5, 1\}$ in Eq.\ref{eq: final_loss}, where $\lambda=0$ corresponds to training without it. As shown in Figure~\ref{fig: lambda_tuning}, the model achieved the highest average Dice score across the three subregions at $\lambda=0.1$. Notably, most positive $\lambda$ values resulted in improvements over the baseline $\lambda = 0$, underscoring the \emph{stability} and \emph{effectiveness} of \rankinglossabbr. This improvement may be attributed to the calibration nature of the loss, which enables \modelabbr to adaptively distinguish between modality-rich and modality-poor conditions, guiding the network to assess each modality’s contribution across various scenarios. As a result, the model learns a more casual, robust relationship between diverse inputs and the task output. In the following, we will verify how $\Loss_{MoFe}$ can calibrate the network to achieve the goal.
\begin{table}[t]
    \centering
    \small
    \setlength{\tabcolsep}{3pt}
    \resizebox{0.9\columnwidth}{!}{%
    \begin{tabular}{l|ccc|ccc}
        \toprule
        & \multicolumn{3}{c|}{\textbf{ECE (\%) $\downarrow$}} & \multicolumn{3}{c}{\textbf{SCE (\%) $\downarrow$}} \\
        \midrule
        Method & ET & TC & WT& ET & TC & WT \\
        \midrule
        MoMKE~\cite{xu2024leveraging} & 3.46 & 4.10 & 4.04 & 14.82 & 11.35 & 7.69  \\
        \modelabbr  & 3.34 & 3.86 & 3.98 & 13.40 & 11.01 & 7.11 \\
         \modelabbr w/ $\Loss_{Conf}$~\cite{ma2023calibrating} & 3.17 & 3.86 & 3.61 & 14.80 & 11.58 & 6.81\\
         \rowcolor{lightgreen}
        \modelabbr w/ $\Loss_{MoFe}$ ( SimMLM) & \textbf{3.15} & \textbf{3.75} & \textbf{3.55} & \textbf{13.21} & \textbf{10.83} &\textbf{5.65} \\
        \bottomrule
    \end{tabular}}
    \caption{Quantitative comparison of \textbf{calibration errors (ECEs$\downarrow$, SCEs$\downarrow$)} averaged across all 15 full and missing modality settings on the validation set of BraTS 2018 dataset.}
    \label{tab: calibration_error}
\end{table}

\vspace{5pt}
\noindent\textbf{\rankinglossabbr leads to better calibration:} To evaluate the performance of $\Loss_{MoFe}$ for model calibration, we compute the \textbf{expected calibration error (ECE) $\downarrow$} and \textbf{static calibration error (SCE) $\downarrow$}~\cite{nixon2019measuring}, both using 20 bins, and compare our method to the most competitive method: MoMKE~\cite{xu2024leveraging}, in addition to the state-of-the-art multimodal calibration approaches, leveraging missing modality information~\cite{ma2023calibrating}, where the task losses $\Loss_{task}$ in $\Loss_{MoFe}$ have been replaced with confidence scores obtained from model predictions using max probability: $\Loss_{Conf}= \max \left( 0, Conf(\bfo^-) - Conf(\bfo^+) \right)$. Results in Table~\ref{tab: calibration_error} are scores directly computed on our validation set, as BraTS does not release ground truth labels on the evaluation set. All experiments were conducted in the same settings.  It is evident that our method with $\Loss_{MoFe}$ achieves the \emph{best calibration performance in all settings, with the lowest ECEs and SCEs}. In the context of this segmentation tasks, compared to $\Loss_{Conf}$, which is label-ignorant and performs comparisons at a low level (e.g., voxels), $\Loss_{MoFe}$ operates at the sample level, and leverages task-specific loss with ground truth information, making it more task-informative. This encourages the network to focus on task-relevant areas across varying levels of missing data, reducing sensitivity to background noise, augmentation artifacts, and outliers.

\begin{table}[t]
    \centering
    \small
    \setlength{\tabcolsep}{10pt}
    \resizebox{0.9\columnwidth}{!}{%
     \begin{tabular}{l|ccc}
        \toprule
        & \multicolumn{3}{c}{\textbf{Counterintuitive rate (\%) $\downarrow$}} \\
        \midrule
        Method & ET & TC & WT \\
        \midrule
        MoMKE~\cite{xu2024leveraging} & 14.60 & 32.47 & 9.86 \\
        \modelabbr w/o $\Loss_{MoFe}$ & 10.47 & 30.94 & 3.87 \\
        \rowcolor{lightgreen}
        \modelabbr w/ $\Loss_{MoFe}$ (SimMLM) & \textbf{7.20} & \textbf{28.19} & \textbf{3.75} \\
        \bottomrule
    \end{tabular}
    }
    \caption{Quantitative comparison of \textbf{counterintuitive rate $\downarrow$} on the BraTS 2018 official evaluation set.}
    \label{tab: CR}
\end{table}

\vspace{5pt}
\noindent\textbf{\rankinglossabbr reduces model counterintuitive rate (CR):} To directly quantify whether $\Loss_{MoFe}$ can better align the model performance with the modality availability (i.e., reduces cases where the model performs better even with fewer input modalities), we introduce a novel metric called the Counterintuitive Rate (CR), which is defined as
\[ \text{CR} = \frac{1}{N} \sum_{i=1}^{N} \frac{1}{|\mathcal{S}_i|} \sum_{(\x_i^-, \x_i^+) \in \mathcal{S}_i} 
    \begin{cases} 
        1 & \text{if} \ P(\x_i^-) > P(\x_i^+) \\
        0 & \text{otherwise}
    \end{cases}.
\]
Here  $N$ is the total number of samples, $\mathcal{S}_i$ is the set of all more vs. fewer modality pairs $(\x_i^+ \; vs.\; \x_i^-)$ for sample $i$. $P(\x_i)$ denotes the model accuracy score (e.g., Dice score for the segmentation task) on the sample $\x_i$. An instance with $P(\x_i^-)>P(\x_i^+)$ is considered as counterintuitive.

We compared CR on BraTS 2018 across three settings: MoMKE~\cite{xu2024leveraging} (previous SOTA), \modelabbr without $\Loss_{MoFe}$, and \modelabbr with $\Loss_{MoFe}$ (SimMLM) and report the performance in Table~\ref{tab: CR}. Results show that  $\Loss_{MoFe}$ achieves the lowest CR, achieving 50.68\%, 13.18\%, and 61.97\% reduction for ET, TC, and WT segmentation, respectively, when compared to MoMKE. Notably, \modelabbr alone also reduces CR significantly, thanks to the dynamic weighting mechanism to automatically identify and prioritize more relevant modalities.

\vspace{5pt}
\noindent\textbf{Broader Impact:} We have shown that SimMLM can provide direct importance values for each input modality and generate informative confidence maps for further analysis in the presence of incomplete data (Fig. \ref{fig: case_study}). Such properties are particularly important to alert clinicians for further investigation in low-resource settings where certain modalities may be unavailable. Additionally, SimMLM's potential for handling varied levels of training-time missing modalities on avMNIST is also discussed in the Supplementary.

\section{Conclusion}
In this paper, we introduced SimMLM, a simple, robust, and flexible framework for multimodal learning in scenarios with test-time missing modalities. By incorporating the Dynamic Mixture of Model Experts (DMoME) architecture and the novel More vs. Fewer (MoFe) ranking loss, SimMLM effectively addresses modality absence. Experimental results demonstrate that SimMLM achieves state-of-the-art performance across both segmentation and classification tasks, excelling in accuracy, efficiency, interpretability, robustness, and reliability. As a generic approach, future work will focus on extending SimMLM to other high-stakes real-world applications where trustworthiness matters and broader scenarios, including training-time missingness.

\newpage

\section*{Acknowledgements}
This research was supported by the UKRI funding (MR/Z505754/1), UK. Dr. Chen is supported by 
Royal Society (RGS/R2/242355).

{
    \small
    \bibliographystyle{ieeenat_fullname}
    \bibliography{main}
}

\clearpage
\clearpage
\setcounter{page}{1}
\maketitlesupplementary
\appendix
\makeatletter
\renewcommand \thesection{A\@arabic\c@section}
\renewcommand\thetable{A\@arabic\c@table}
\renewcommand \thefigure{A\@arabic\c@figure}
\makeatother

\paragraph{Implementation of the gating network }

The gating network $G$ in our DMoME framework is designed with a generic and efficient architecture to ensure adaptability across diverse tasks and modalities. At a high level, it consists of a feature extraction block (e.g., 2D or 3D convolutional neural networks (CNNs), transformers, etc.), followed by a linear layer that generates low-dimensional weight vectors. This architecture strikes a balance between simplicity and flexibility, enabling it to adjust modality contributions effectively across tasks. Furthermore, its modular design allows seamless accommodation of varying task requirements and diverse input modalities, ensuring broad applicability and adaptability.

As shown in Table~\ref{sup_fig: gating_design}, for the BraTS 2018 segmentation task, the input modalities are four input imaging modalities (\(4 \times 128 \times 128 \times 128\) with zeros if certain modalities are missing). Here, since they are all images sharing the same dimensionality, we simply extract \textbf{$256-dim$ }features using a 5-layer 3D CNN to directly fuse visual information and send the features to an linear layer to get a  \(4 \times 3\) gating vector (\# modalities $\times$ \#tasks). The vector will be transformed using the softmax function to generate gating weights, which adjust the contribution of each modality per task. In contrast, the avMNIST classification task involves heterogeneous modalities: the image modality (\(1 \times 28 \times 28\)) and the audio modality  (\(1 \times 20 \times 20\)). To account for this heterogeneity, two separate 2-layer CNNs are used to extract 128-dimensional feature vectors for each modality. These features are concatenated to form the same size of \textbf{$256-dim$ } representation, which is then processed by a similar linear layer to generate a \(2 \times 1\) gating vector. This modular approach ensures flexibility and robust feature integration across diverse data types. 

In this work, we simply designed these two lightweight gating networks empirically, which provides substantial improvements. Thanks to the flexibility of the framework, readers are encouraged to explore more advanced architectures for their customized tasks.

\begin{table}[ht]
\centering
\footnotesize
\setlength{\tabcolsep}{3pt}
\resizebox{0.9\columnwidth}{!}{%
\begin{tabular}{c|c|cc}
\hline
\textbf{General design} & BraTS 2018 & \multicolumn{2}{c}{avMNIST} \\ \hline
\multirow{4}{*}{\textbf{Feature extractor}} & \begin{tabular}[c]{@{}c@{}}4-modality MRIs\\ \(4 \times 128 ^3\)\end{tabular} & \multicolumn{1}{c|}{\begin{tabular}[c]{@{}c@{}}image \\ \(1 \times 28^2 \)\end{tabular}} & \begin{tabular}[c]{@{}c@{}}audio \\ \(1 \times 20^2\)\end{tabular} \\ \cline{2-4} 
 & \begin{tabular}[c]{@{}c@{}}5-layer \\ CNNs (3D)\end{tabular} & \multicolumn{1}{c|}{\begin{tabular}[c]{@{}c@{}}2-layer \\ CNNs (2D)\end{tabular}} & \begin{tabular}[c]{@{}c@{}}2-layer \\ CNNs (2D)\end{tabular} \\ \cline{2-4} 
 & Flatten & \multicolumn{2}{c}{Flatten \& concat} \\ \cline{2-4} 
 & 256-dim feature & \multicolumn{2}{c}{256-dim feature} \\ \hline
\textbf{Linear layer} & \begin{tabular}[c]{@{}c@{}}Linear layer \\ Output size: \(4 \times 3\)\\ (4 modalities $\times$ 3 tasks)\end{tabular} & \multicolumn{2}{c}{\begin{tabular}[c]{@{}c@{}}Linear layer \\ Output size: 2 $\times$ 1 \\ (2 modalities $\times$ 1 task)\end{tabular}} \\ \hline
\end{tabular}%
}
\caption{Designs of the gating network $G$ for the BraTS 2018 segmentation task and the avMNIST classification task.}
\label{sup_fig: gating_design}
\end{table}

\paragraph{Importance of the design choice and the training recipe in SimMLM}
\label{app: dmome_experiments}
\begin{table}[ht]
    \centering
    \label{tab:dmome_modular_ablation}
    \resizebox{0.9\columnwidth}{!}{%
    \begin{tabular}{lccc}
        \toprule
        & \multicolumn{3}{c}{\textbf{Dice score $\uparrow$}} \\
        \midrule
        Method & ET & TC & WT \\
        \midrule
        DMoME w/o expert pretraining & 62.24 & 77.42 & 86.63 \\
        DMoME w/ static averaging & 61.40 & 77.49 & 86.70 \\
        \textbf{DMoME w/ dynamic weighting (SimMLM)} & \textbf{63.22} & \textbf{78.54} & \textbf{87.21} \\
        \bottomrule
    \end{tabular}%
    }
    \caption{Importance of the design choice and the training recipe in SimMLM. Reported values are the average segmentation performance on the BraTS 2018 validation set, across all missing \& full modality settings.}
    \label{sup_tab: dmome_modular_ablation}
\end{table}

In this section, we would like to highlight the contribution of pretraining and dynamic weighting by comparing our proposed DMoME with its two variants: 
\begin{itemize}
    \item \textbf{DMoME w/o expert pretraining}: Here, all modality experts and the gating network $G$ were directly co-trained from random initialization, without pretraining. The total number of training epochs is the same as the proposed two-stage one for fair comparison.  
    \item \textbf{DMoME w/ static averaging}: In this setting, the gating network $G$ is replaced with a simple static averaging operation. The final output can be simply viewed as the mean of the available modality experts' outputs.
\end{itemize}
Results in Table~\ref{sup_tab: dmome_modular_ablation} demonstrate that both expert pretraining and dynamic reweighting contribute to the success of SimMLM. Skipping expert pretraining (\textbf{DMoME w/o expert pretraining}) results in a noticeable performance drop, highlighting the critical role of independent pretraining. This step enables modality experts to focus on task-relevant knowledge without interference from cross-expert interactions, which could be noisy especially at the beginning of training. Replacing the dynamic gating mechanism with static averaging (\textbf{DMoME w/ static averaging}) also causes a significant performance drop across all subregions (ET, TC, WT). This underscores the effectiveness of dynamically assigning weights to modality experts, which is not just a simple model ensembling strategy.  

\begin{figure}[ht]
    \centering
    \includegraphics[width=0.9\columnwidth]{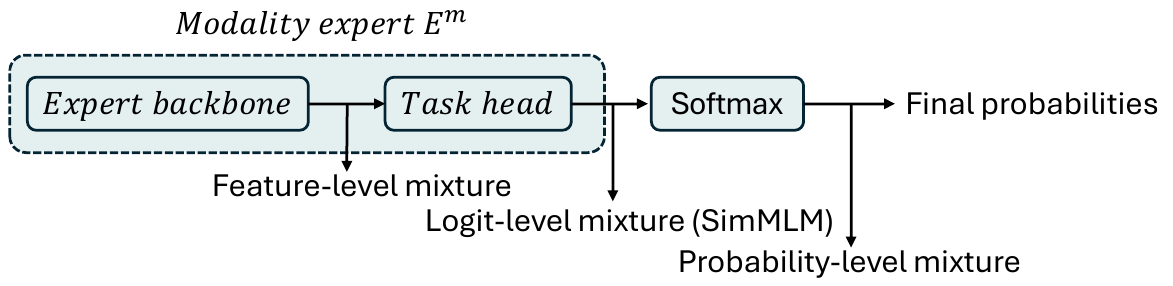} 
    \caption{Three types of weighted sum micro design: feature-level, logit-level (SimMLM), and probability-level.}
    \label{sup_fig: gating_micro_design_cropped}
\end{figure}

\paragraph{Impact of performing dynamic weighting at different levels in DMoME}
As shown in Figure~\ref{sup_fig: gating_micro_design_cropped}, we propose a simple yet effective mixture of modality experts strategy, which directly applies expert weighting at the logit level (before softmax) instead of probabilities. 

Here, we tested the importance of this design choice, by comparing it to the other two alternatives with the same gating network design:
\begin{itemize}
    \item \textbf{Feature-level mixture}: Here, the dynamic weighting is applied at hidden feature maps (before passing through the segmentation/classification head) rather than the output space, which is adopted in MoMKE~\cite{xu2024leveraging}. The feature maps from modality experts are later mixed with the weights produced by the gating network. A segmentation head is followed to generate the final results given the mixed feature map.
    \item \textbf{Probability-level mixture}: Here the dynamic weighting is applied to the probabilities from each expert (after softmax).
\end{itemize}
Results in Table~\ref{sup_tab: weighted_sum_ablation} show that performing logit-level mixture generally achieves better results, especially on the more challenging enhancing tumor (ET) and tumor core (TC) segmentation tasks. We believe this improvement stems from SimMLM's logit-level weighting, where the weighting parameter functions as a temperature to rescale the confidence of each modality’s prediction in the final output. This approach aligns with the principles of temperature rescaling~\cite{guo2017calibration} for model calibration, which can effectively adapt model predictions to better reflect the likelihood of ground truth correctness, leading to more reliable multi-modal inference especially in challenging scenarios. In our main paper, we have already demonstrated that our DMoME framework, when used standalone, can achieve much lower calibration errors compared to MoMKE in Table~5, thanks to the logit-level dynamic weighting. 
\begin{table}[ht]
    \centering
    \label{tab:weighted_sum_ablation}
    \resizebox{0.9\columnwidth}{!}{%
    \begin{tabular}{lccc}
        \toprule
        & \multicolumn{3}{c}{\textbf{Dice score $\uparrow$}} \\
        \midrule
        Method & ET & TC & WT \\
        \midrule
        Feature-level mixture & 62.76 & 78.19 & 87.27 \\
        Probability-level mixture & 62.72 & 78.31 & \textbf{87.33} \\
        \textbf{Logit-level mixture (SimMLM)} & \textbf{63.22} & \textbf{78.54} & 87.21 \\
        \bottomrule
    \end{tabular}%
    }
    \caption{Importance of performing dynamic weighting at logit level. Reported values are the average segmentation performance on the BraTS 2018 validation set, across all missing \& full modality settings.}
    \label{sup_tab: weighted_sum_ablation}
\end{table}

\begin{figure}[t]
    \centering\includegraphics[width=0.8\columnwidth]{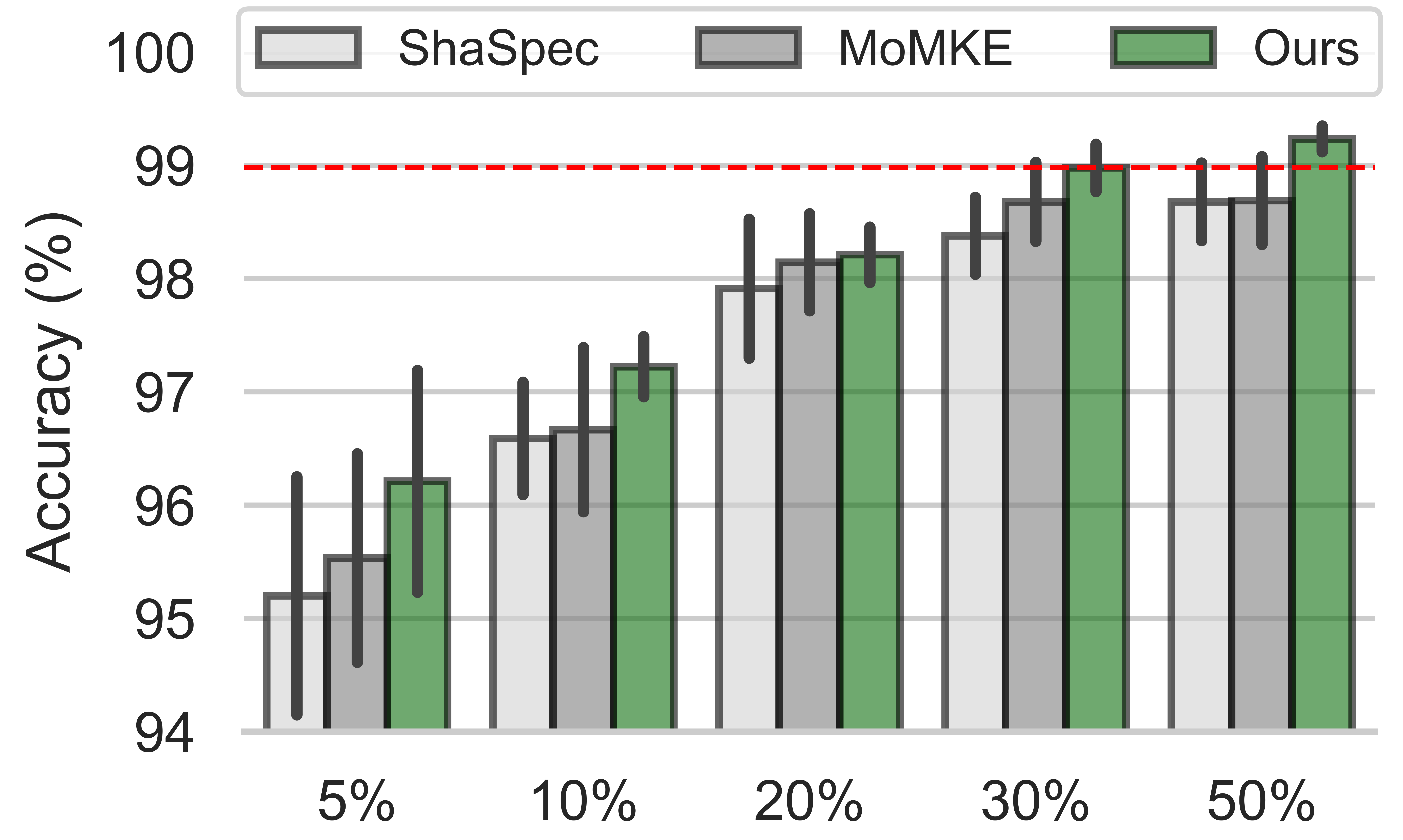} 
    \caption{
    Model performance comparison with varying rates of missing audio data during training on av-MNIST.
    }
    \label{fig: training_missing}
\end{figure}

\begin{figure}[t]
    \centering
    \includegraphics[width=0.88\linewidth]{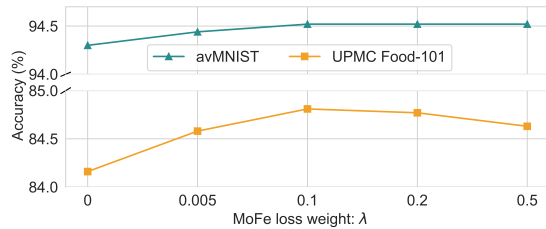}
    \caption{MoFe weight $\lambda$ tuning on avMNIST, UPMC Food-101.}

    \label{fig: lambda_tuning_cls}
\end{figure}

\paragraph{SimMLM's robustness against varied levels of missing modalities at training time} To further investigate SimMLM's resilience to training-time missingness, we conducted experiments where the audio coverage rate was reduced to {5\%, 10\%, 20\%, 30\%, 50\%} during the training. We then evaluated the model on the full modality evaluation set and compared its performance to ShaSpec~\cite{wang2023multi} and MoMKE~\cite{xu2024leveraging}. As shown in Figure~\ref{fig: training_missing}, our model consistently achieves the highest accuracy in all settings, demonstrating substantial improvement in handling training-time modality absence. Remarkably, our model achieves even better accuracy comparable to competing methods that use 50\% of the audio data, with only 30\% of it during training (see red dashed line). These results demonstrate our method's potential to handle training-time modality missingness as well.

\paragraph{MoFe coefficient tuning on avMNIST and UPMC Food-101} 
Setting the MoFe loss coefficient to 0.1 generalizes well across tasks. As shown in Fig.~\ref{fig: lambda_tuning_cls}, where MoFe consistently improves performance across a range of values, with 0.1 yielding stable results. 

\paragraph{Discussion: How well does the unimodal performance align with the importance derived from the learned gating weights?}
Interestingly, the gating weights show weak to moderate alignment with unimodal performance \emph{at the sample level}. On UPMC Food-101, the Pearson correlation between each modality’s gating weight and its unimodal confidence on the correct class is 0.29 for image and 0.41 for text. This partial alignment suggests that while unimodal confidence reflects individual modality strength, it may not fully capture cross-modal complementarity or sample difficulty—both of which influence the model’s reliance during inference. We consider this behavior desirable, as it indicates the gating mechanism is context-aware rather than simply mirroring unimodal confidence.

\noindent\paragraph{Clinical value of SimMLM} In real-world clinical scenarios, especially in low-resource settings, certain imaging modalities may be unavailable due to technical issues, equipment limitations, or patient factors. SimMLM’s ability to generate more reliable and accurate segmentation under these conditions is essential. The model not only provides direct importance values for each input modality but also generates more informative confidence maps, including entropy maps, for voxel-wise predictions in the output space. This capability is particularly valuable in critical applications like tumor detection and diagnosis, where accurate and detailed segmentation is crucial for treatment planning and patient care. By enabling clinicians to make confident decisions—whether by providing insights or suggesting further investigation or special care for uncertain regions—even in the presence of incomplete data, SimMLM greatly enhances the robustness and reliability of clinical workflows. In the future, we will explore SimMLM's application in broader, high-stakes scenarios, such as surgical scene reconstruction, action segmentation, and surgical planning. 

\end{document}